%% file: main.tex
\documentclass{article} % For LaTeX2e
\usepackage{iclr2025_conference,times}

\usepackage{wrapfig}
\usepackage{lipsum}

\input{math_commands.tex}

\usepackage{booktabs}
\usepackage{multirow}
\usepackage{amssymb}
\usepackage{graphicx}

\usepackage{colortbl}
\usepackage{xcolor}

\usepackage{hyperref}
\usepackage{url}
\usepackage{multicol}
\usepackage{aliascnt}
\usepackage{adjustbox}
\usepackage{siunitx}
\usepackage{enumitem}
\usepackage{amsmath}
\usepackage{longtable}
\usepackage{subcaption}
\usepackage{calc}
\usepackage[most]{tcolorbox}
\usepackage{listings}

\usepackage{microtype}
\usepackage{mathtools}
\usepackage{amsthm}
\usepackage{algorithm}
\usepackage{algorithmic}

\definecolor{uclablue}{rgb}{0.15, 0.45, 0.68}
\hypersetup{
    breaklinks,
    citecolor=uclablue,
    colorlinks=true,
}

\newtcolorbox{AIbox}[2][]{aibox,title=#2,#1}
\tcbset{
  aibox/.style={
    top=10pt,
    colframe=black,
    colbacktitle=black,
    coltitle=white,
    center,
  }
}

\lstdefinelanguage{prompt}{
    basicstyle=\scriptsize\ttfamily, 
    mathescape=true,
    escapebegin=\color{latentcolor},
    escapeend={},
    escapechar=@,
    stringstyle = \color{myorange},
    showstringspaces = false,
    moredelim = [s][\color{mypink}]{`}{`},
    moredelim = [s][\color{mybrown}]{```json}{```},
    moredelim = [s][\color{latentcolor}]{<StartOfLatent>}{<EndOfLatent>},
    literate = %
        {\ \ a.\ }{{\textcolor{mypurple}{\ \ a.\ }}}5
        {\ \ b.\ }{{\textcolor{mypurple}{\ \ b.\ }}}5
        {\ \ c.\ }{{\textcolor{mypurple}{\ \ c.\ }}}5
        {\ \ d.\ }{{\textcolor{mypurple}{\ \ d.\ }}}5
        {\ \ e.\ }{{\textcolor{mypurple}{\ \ e.\ }}}5
        {\ \ f.\ }{{\textcolor{mypurple}{\ \ f.\ }}}5
        {\ \ g.\ }{{\textcolor{mypurple}{\ \ g.\ }}}5
        {\ \ h.\ }{{\textcolor{mypurple}{\ \ h.\ }}}5
        {\ I.\ }{{\textcolor{mypurple}{\ I.\ }}}4
        {\ II.\ }{{\textcolor{mypurple}{\ II.\ }}}5
        {\ III.\ }{{\textcolor{mypurple}{\ III.\ }}}6
        {\ IV.\ }{{\textcolor{mypurple}{\ IV.\ }}}5
        {\ V.\ }{{\textcolor{mypurple}{\ V.\ }}}4
}

\newtcblisting[auto counter, number within=subsection]{prompt}[4][]{%
  before skip=6pt, after skip=6pt,
  top=6pt, bottom=6pt, left=7pt, right=7pt,
  enhanced, breakable,
  colback=white, colframe=gray!65,
  boxrule=0.6pt, arc=2pt,
  drop shadow={black!15},
  listing only,
  listing options={
    language=prompt,
    upquote=true,
    basicstyle=\scriptsize\ttfamily \setlength{\baselineskip}{1.1\baselineskip},
    columns=fullflexible,
    breaklines=true,
    breakindent=0pt,
    xleftmargin=\ifx\empty#2\empty0pt\else#2\fi,
    xrightmargin=\ifx\empty#3\empty0pt\else#3\fi,
    aboveskip=0pt,
    belowskip=0pt,
    showstringspaces=false,
  },
  title={\ifstrempty{#4}{Prompt \thetcbcounter}{Prompt \thetcbcounter: #4}},
  colbacktitle=gray!6, coltitle=black,
  attach boxed title to top left={yshift=-1mm, xshift=6pt},
  boxed title style={colback=gray!6, boxrule=0pt, sharp corners},
  #1
}

\newtcblisting[auto counter, number within=subsection]{showcase}[4][]{%
  before=\par\vspace{\baselineskip},
  after=\par,
  width=\linewidth,
  enhanced,
  arc=0em,
  boxrule=1pt,
  listing only,
  listing options={
    language=prompt,
    upquote=true,
    basicstyle=\scriptsize\ttfamily \setlength{\baselineskip}{1.1\baselineskip},
    breaklines=true,
    breakindent=0pt,
    xleftmargin=\ifx\empty#2\empty-12pt\else#2\fi,
    xrightmargin=\ifx\empty#3\empty-5pt\else#3\fi,
    aboveskip=-4pt,
    belowskip=-4pt,
    columns=fullflexible,
    },
  colback=white,
  colframe=gray,
  colbacktitle=gray!5,
  coltitle=black,
  attach boxed title to top center={yshift=-3mm},
  box align=center,
  parbox=false,
  title={\ifstrempty{#4}{Example \thetcbcounter}{Example \thetcbcounter: #4}},
  #1
}

\definecolor{linkColor}{rgb}{0.2,0.4,0.6}
\definecolor{myblue}{HTML}{0379AC}
\definecolor{myred}{HTML}{A50E50}
\definecolor{myorange}{RGB}{238, 133, 74}
\definecolor{latentcolor}{named}{cyan}
\definecolor{normalcolor}{RGB}{0, 0, 0}
\usepackage{marvosym}

\theoremstyle{plain}

\theoremstyle{definition}

\theoremstyle{remark}

\newcommand*\samethanks[1][\value{footnote}]{\footnotemark[#1]}

\title{PolicyLong: Towards On-Policy Context Extension}

\author{
Junlong Jia\thanks{Equal contribution. This paper is published as a conference paper at COLM 2026.} \quad
Jiang Zhou\samethanks \quad
Ziyang Chen\samethanks \quad
Xing Wu\thanks{Corresponding author. Correspondence to \{ucaswu,maxwellyu\}@tencent.com.} \quad
Chaochen Gao \\
TingHao Yu\samethanks \quad
Feng Zhang \quad
Songlin Hu \\[0.5em]
Hunyuan Team, Tencent \quad \quad Chinese Academy of Sciences
}

\begin{document}
\maketitle

\begin{abstract}
Extending LLM context windows is hindered by scarce high-quality long-context data. Recent methods synthesize data with genuine long-range dependencies via information-theoretic verification, selecting contexts that reduce a base model's predictive entropy. However, their single-pass offline construction with a fixed model creates a fundamental \emph{off-policy gap}: the static screening landscape misaligns with the model's evolving capabilities, causing the training distribution to drift. We propose \textbf{PolicyLong}, shifting data construction towards a dynamic on-policy paradigm. By iteratively re-executing data screening (entropy computation, retrieval, and verification) using the \emph{current} model, PolicyLong ensures the training distribution tracks evolving capabilities, yielding an emergent self-curriculum. Crucially, both positive and hard negative contexts derive from the current model's entropy landscape, co-evolving what the model learns to exploit and resist. Experiments on RULER, HELMET, and LongBench-v2 (Qwen2.5-3B) show PolicyLong consistently outperforms EntropyLong and NExtLong, with gains growing at longer contexts (e.g., +2.54 at 128K on RULER). The improvements further hold when scaling up to a 7B model (OLMo3-7B), confirming that the value of on-policy data evolution generalizes across model family and scale.
\end{abstract}

\section{Introduction}
\label{sec:intro}

Processing long contexts effectively is currently a core challenge for large language models \citep{huo2025dots,liu2024deepseek}. Although architectural advances such as rotary position embeddings \citep{su2021roformer} and efficient attention \citep{dao2022flashattention} have extended context windows beyond 128K tokens, high-quality long-context training data remain scarce and are a critical bottleneck. Since such documents are rare in web-scale corpora, recent methods \citep{gao2024quest,nextlong2025,jia2025entropylong} instead synthesize training data by concatenating short documents and verifying genuine long-range dependencies \citep{fang2024wrong}. For instance, EntropyLong \citep{jia2025entropylong} uses information-theoretic verification to measure whether prepending a candidate context reduces the base model's predictive entropy at high-uncertainty positions, thereby certifying meaningful cross-document dependencies.

Despite their principled design, previous methods share a fundamental limitation: they construct all training data in a single offline pass using a fixed base model, introducing an \emph{off-policy gap}. As training progresses, the model's capabilities evolve: positions that were initially uncertain become trivial and provide diminishing optimization signal. In other words, the static dataset, calibrated to the base model's initial uncertainty profile, grows increasingly misaligned with the trained model's actual information deficits, yielding diminishing returns. This observation suggests that long-context data construction should not be a static preprocessing step, but a \emph{closed loop} that co-evolves with model training.

Recent advances in on-policy learning and self-distillation \citep{hubotter2026reinforcement, zhao2026self, qu2026pope, shenfeld2026self} likewise show that models learn more effectively when supervision is derived from their current policy, yet this principle remains largely unexplored in long-context data synthesis, which still relies exclusively on fixed base models. Inspired by these advances, we propose \textbf{PolicyLong}, a framework that shifts long-context data construction to a dynamic, on-policy paradigm. At its core lies an \emph{Iterative Self-Curriculum} (\S\ref{sec:curriculum}): we partition the training process into $T$ stages, where the \emph{current} model $M_t$ re-executes the complete data screening pipeline---entropy computation, retrieval, and verification---at each stage. As $M_t$ improves, its entropy landscape shifts: previously challenging positions become low-entropy (mastered), while the remaining high-entropy positions represent genuinely difficult long-range dependencies. This yields an implicit curriculum without explicit difficulty scheduling: early stages surface relatively basic dependencies, while later stages naturally concentrate on the model's true learning frontier.

\begin{figure}[t]
    \centering
    \includegraphics[width=\linewidth]{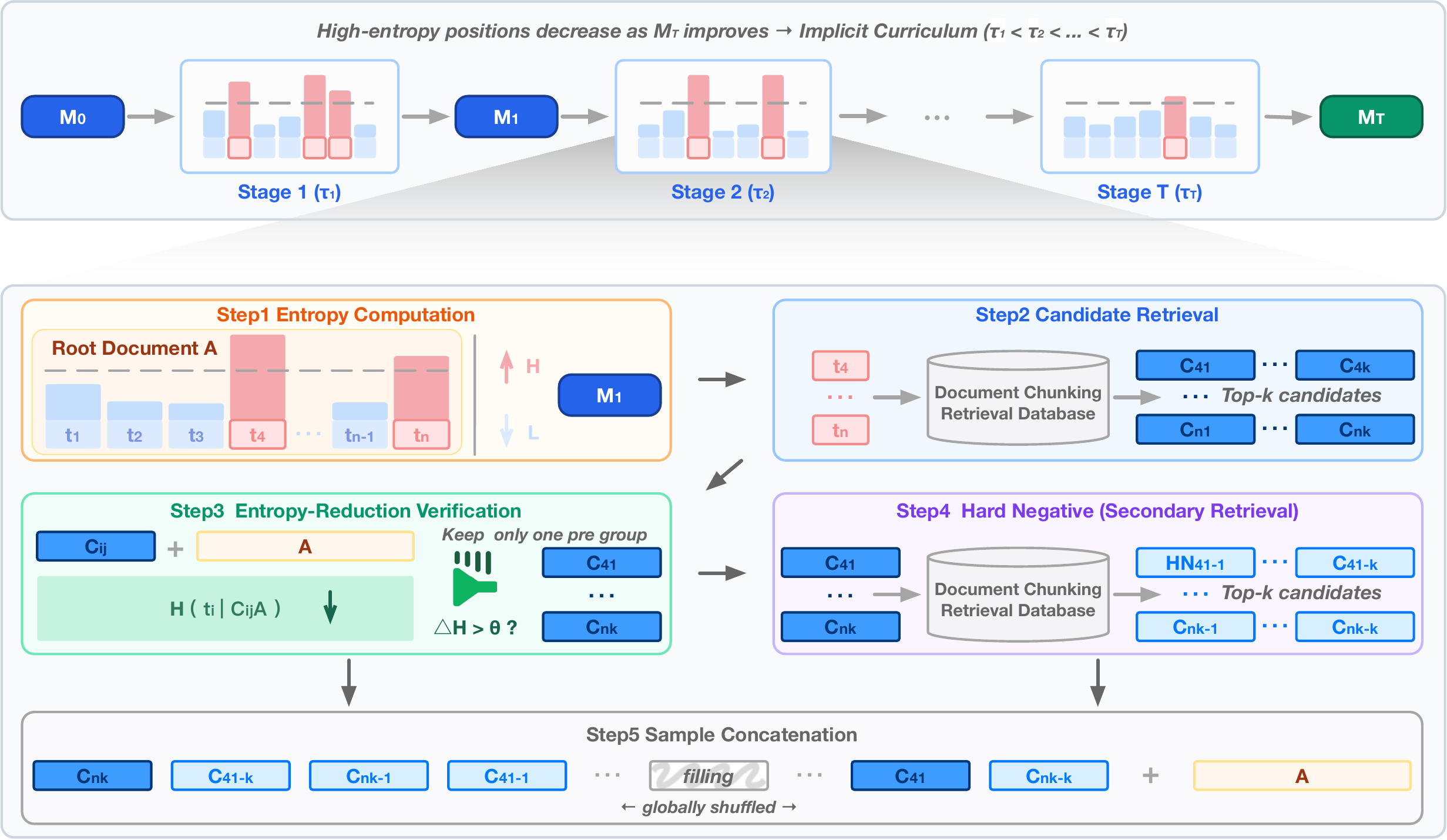}
    \vspace{-2mm}
    \caption{Overview of PolicyLong. Top: iterative on-policy training across stages, where the current model $M_t$ re-executes data screening and progressively shifts training toward harder dependencies, yielding an implicit self-curriculum. Bottom: per-stage data construction, including entropy computation on root document $A$, candidate retrieval, entropy-based positive verification, on-policy hard negative construction, and final sequence assembly.}
    \label{fig:policylong_overview}
    \vspace{-4mm}
\end{figure}

Crucially, this on-policy principle extends beyond positive context selection. At each stage, we retrieve candidates that are semantically similar to the identified positive contexts, treated as hard negatives (\S\ref{sec:hardneg}). Because these negatives are tied to the current model's uncertainty, their difficulty naturally aligns with the model's capabilities: as the model learns to handle long-context dependencies better, the hard negatives it encounters shift accordingly. Consequently, the entire training distribution---dictating both what the model should exploit and what it must resist---is governed by a unified on-policy mechanism, overviewed in Figure~\ref{fig:policylong_overview}.

Experiments on RULER \citep{hsieh2024ruler}, HELMET \citep{yen2024helmet}, and LongBench-v2 \citep{bai2024longbench2} with Qwen2.5-3B demonstrate that PolicyLong consistently outperforms strong baselines NExtLong \citep{nextlong2025} and EntropyLong \citep{jia2025entropylong}, with gains that grow with context length (e.g., +2.54 at 128K on RULER). We further validate the method on OLMo3-7B \citep{olmo2025olmo} under a production-scale 50B-token budget, confirming that the advantage carries across model family (Qwen2.5 $\to$ OLMo3) and scale (3B $\to$ 7B). In addition, we provide a comprehensive analysis of the key design choices of PolicyLong to understand its effectiveness.

Our contributions are summarized as follows:

\begin{enumerate}[itemsep=0.2em,parsep=0pt,topsep=0.5em]
    \item We propose \textbf{PolicyLong}, an on-policy long-context training framework that iteratively refreshes data with the \emph{current} model, yielding an implicit self-curriculum as the model's learning frontier evolves.
    \item We introduce a unified on-policy data construction strategy for both positive contexts and hard negatives, enabling the model to exploit useful distant evidence while resisting plausible distractors.
    \item Experiments on RULER, HELMET, and LongBench-v2 show that PolicyLong achieves the strongest overall long-context performance, especially at longer contexts, and that its advantage holds across model family and scale.

\end{enumerate}

\section{Preliminary Analysis: Evidence for the Off-Policy Gap}
\label{sec:prelim_exp}

Before presenting our method, we provide empirical evidence for the two core motivations behind PolicyLong: (1)~static base-model data becomes progressively misaligned with the evolving model, and (2)~base-model-identified dependencies become insufficiently challenging as training proceeds.

\begin{figure}[t]
    \centering
    \begin{subfigure}[t]{0.48\textwidth}
        \centering
        \includegraphics[width=\linewidth]{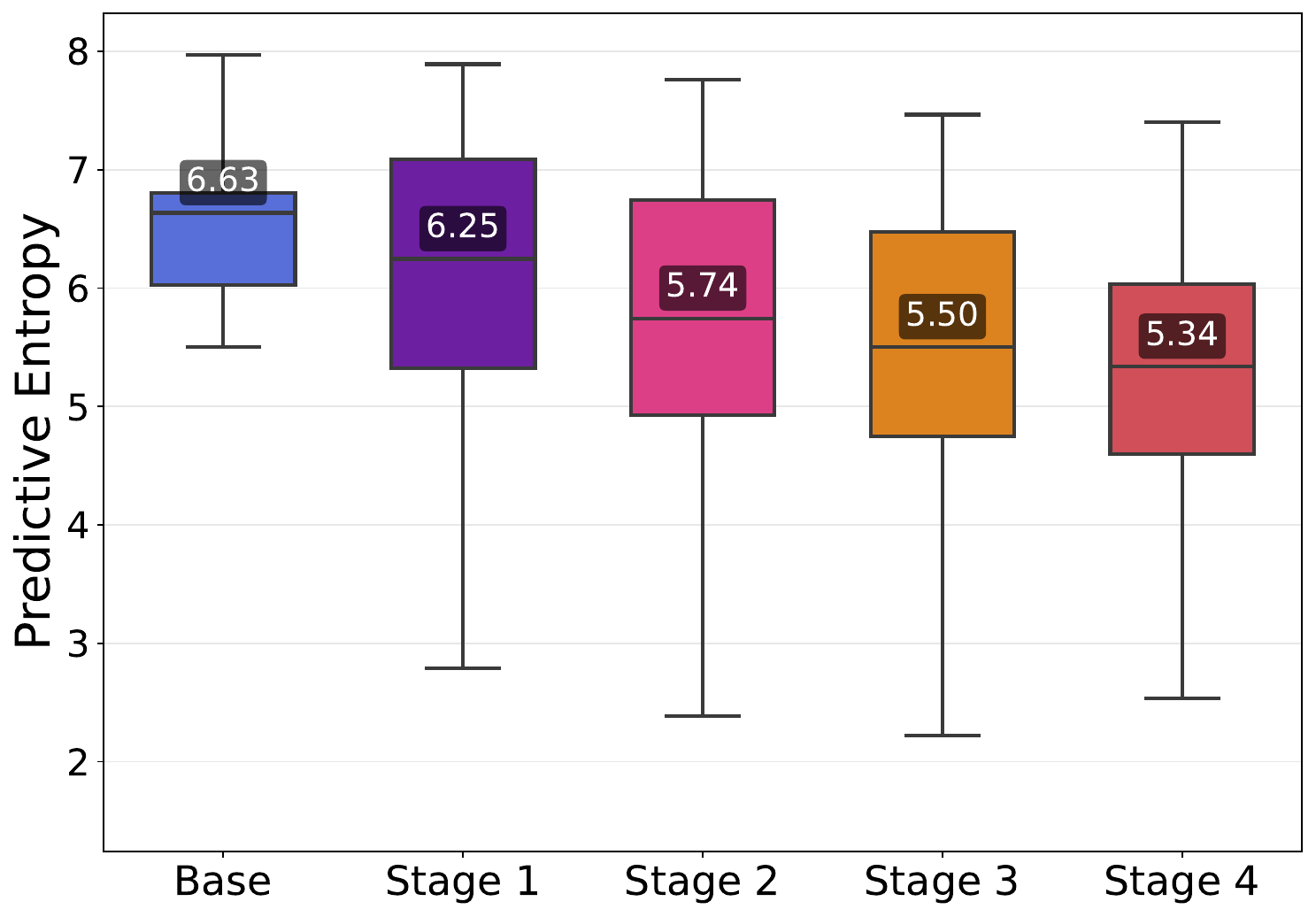}
        \caption{Entropy of base-model high-entropy positions re-evaluated on trained models.}
        \label{fig:entropy_decay}
    \end{subfigure}
    \hfill
    \begin{subfigure}[t]{0.48\textwidth}
        \centering
        \includegraphics[width=\linewidth]{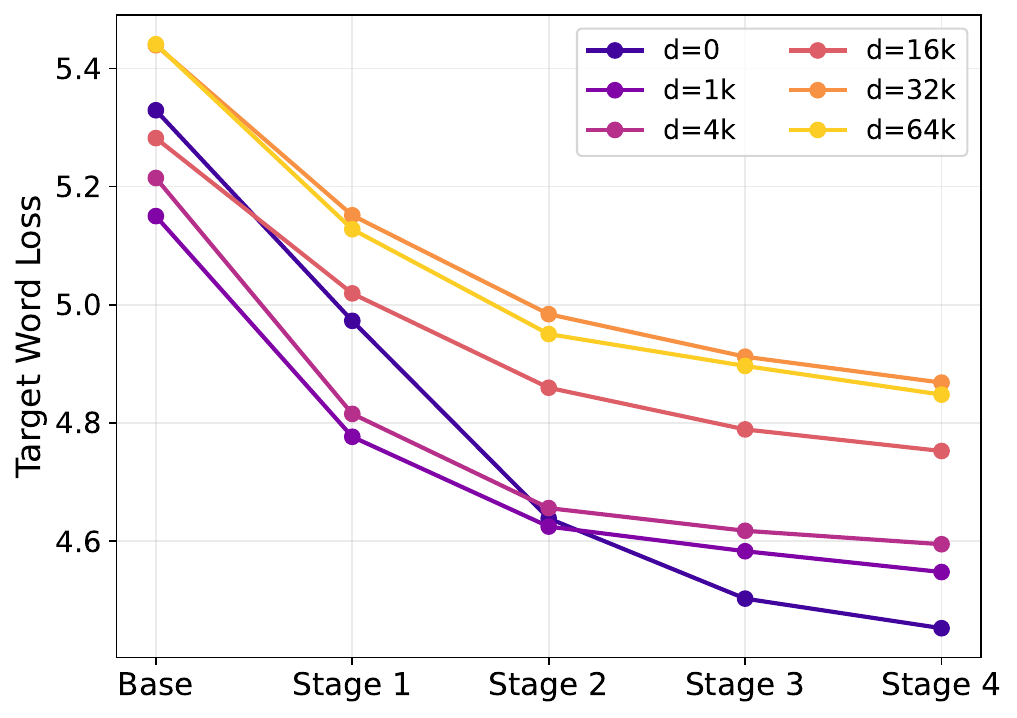}
        \caption{Target-word loss with positive context across stages at varying filler distances $d$.}
        \label{fig:loss_with_chunk}
    \end{subfigure}
    \caption{Empirical evidence for the off-policy gap. (a)~The entropy distribution of positions initially identified as high-entropy by the base model shifts significantly across stages, demonstrating progressive data--model misalignment. (b)~Target-word loss with the positive context decreases rapidly, indicating that base-model-identified dependencies become trivially easy for the trained model.}
    \label{fig:offpolicy_evidence}
\end{figure}

\textbf{Static data becomes misaligned with the evolving model.}
We track the predictive entropy of positions originally identified as high-entropy by the fixed base model, re-evaluating them using progressively trained models $M_t$ at each stage. Figure~\ref{fig:entropy_decay} reveals a consistent decline in median entropy from 6.63 to 5.34 across stages, with the interquartile range narrowing substantially. Rather than remaining uniformly high-entropy, many of these positions are resolved by the evolving model, while its actual learning frontier moves elsewhere. This means that a static dataset, calibrated solely to the base model's initial uncertainty profile, fails to capture the model's shifting knowledge gaps, providing direct evidence for the off-policy gap.

\textbf{Base-model dependencies provide diminishing training signal.}
We further track the target-word loss on fixed base-model-selected data (with the verified positive context chunk $B$ prepended) across training stages. Figure~\ref{fig:loss_with_chunk} shows that loss decreases consistently across all filler distances $d \in \{0, 1\text{k}, 4\text{k}, 16\text{k}, 32\text{k}, 64\text{k}\}$, with the steepest decline at shorter distances. This indicates that the model rapidly masters the base-model-identified dependencies, receiving progressively weaker gradient signals. We further verify in Section~\ref{sec:ablation} (Figure~\ref{fig:data_difficulty}) that data constructed on-policy at later stages exhibits significantly higher difficulty than static data, with the gap widening across stages---confirming that the off-policy gap not only exists but compounds over time. The diminishing challenge from static positive dependencies motivates two design choices: (1)~iterative on-policy data refresh to keep the training distribution aligned with the model's evolving frontier, and (2)~on-policy hard negative construction to maintain training difficulty, both detailed next.

\section{Method}
\label{sec:method}

Figure~\ref{fig:policylong_overview} summarizes the full framework. We first recap the EntropyLong pipeline on which PolicyLong is built (\S\ref{sec:preliminaries}), then present the two key mechanisms that turn it into an on-policy closed loop: iterative self-curriculum (\S\ref{sec:curriculum}) and on-policy hard negative construction (\S\ref{sec:hardneg}).

\subsection{Preliminaries: EntropyLong recap}
\label{sec:preliminaries}

We briefly review the EntropyLong pipeline, which forms the foundation of our approach. Given a root document $A$ and a retrieval corpus $\mathcal{R}$:

\begin{enumerate}[itemsep=0.1em,parsep=0pt,topsep=0.3em]
    \item \textbf{Entropy computation}: Compute per-token entropy $H_{M}(x_i | x_{<i})$ for each position in $A$ using the base model $M$.
    \item \textbf{High-entropy position identification}: Select positions where entropy exceeds a threshold $\tau$.
    \item \textbf{Candidate retrieval}: Use text fragments around high-entropy positions as queries to retrieve top-$k$ candidates from $\mathcal{R}$.
    \item \textbf{Entropy-reduction verification}: For each candidate $B$, verify that prepending $B$ to $A$ significantly reduces the predictive entropy at the identified high-entropy position $x_i$:
    \begin{equation}
    \Delta H(B, x_i) = \frac{H_{M}(x_i | x_{<i}) - H_{M}(x_i | B, x_{<i})}{H_{M}(x_i | x_{<i})} > \tau_{\text{pos}}
    \label{eq:entropy_reduction}
    \end{equation}
    where $x_{<i}$ represents the context preceding $x_i$ within $A$.
    \item \textbf{Sequence construction}: Randomly shuffle all verified positive contexts $\{B_1, B_2, \ldots\}$ and prepend them to $A$: $[\text{shuffle}(B_1, B_2, \ldots)]\;A$.
\end{enumerate}

\subsection{Iterative Self-Curriculum}
\label{sec:curriculum}

The central idea of PolicyLong is to replace EntropyLong's single-stage, off-policy data construction with an iterative, on-policy closed loop. We partition training into $T$ stages. At each stage $t \in \{0, 1, \ldots, T{-}1\}$:

\begin{enumerate}[itemsep=0.1em,parsep=0pt,topsep=0.3em]
    \item Sample a fresh document set $\mathcal{D}_t$ from the source corpus $\mathcal{D}$.
    \item Execute the full dependency screening and construction pipeline (Section~\ref{sec:hardneg}) using the \emph{current} model $M_t$.
    \item Obtain training data $\mathcal{S}_t$ (fixed at $N$ tokens per stage).
    \item Train $M_t$ on $\mathcal{S}_t$ to obtain $M_{t+1}$.
\end{enumerate}

Each stage uses a fixed budget of $N$ newly sampled tokens (e.g., 1B), yielding a total of $N \times T$ training tokens. The crucial difference from EntropyLong is that each stage re-executes the \emph{complete} data screening pipeline with the current on-policy model $M_t$, rather than reusing a static dataset. As $M_t$ improves, its entropy landscape shifts: previously high-entropy positions may become low-entropy (mastered), while remaining high-entropy positions represent the model's true learning frontier.

\textbf{Implicit curriculum.}
This design gives rise to an implicit curriculum without explicit difficulty scheduling. In early stages, the model is weak and many positions exhibit high entropy, yielding relatively basic dependencies. In later stages, only genuinely difficult positions remain high-entropy, and the screening pipeline automatically selects more challenging dependencies. The curriculum arises naturally from the closed loop between the model's evolving uncertainty and the data construction process.

Rather than keeping the entropy threshold $\tau$ fixed, we let it increase across stages, taking progressively larger values over training (e.g., $\tau \in \{0.2, 0.4, 0.6, 0.8\}$ for $T=4$). As the model improves, this rising threshold keeps each stage focused on an increasingly stringent high-entropy frontier, regardless of how the overall entropy distribution shifts.

\subsection{On-Policy Hard Negative Construction}
\label{sec:hardneg}

Hard negatives are effective for robust long-context training \citep{nextlong2025}, and in PolicyLong they arise naturally from the same on-policy screening process used for positive contexts. Because the full pipeline is re-executed with the current model $M_t$ at every stage, their identity and difficulty automatically co-evolve with the model.

For each sampled document in $\mathcal{D}_t$, we perform the following steps.

\noindent\textbf{High-entropy position identification and candidate retrieval.}

Following Section~\ref{sec:preliminaries}, we first use $M_t$ to compute per-token entropy on the root document $A$, select high-entropy positions via the stage-increasing threshold $\tau$, and retrieve top-$k$ candidates from the corpus $\mathcal{R}$ using fragments around those positions as queries.

\noindent\textbf{Hard negative construction via secondary retrieval.}

EntropyLong trains the model to \emph{exploit} useful context, but not to \emph{resist} plausible distractors. To capture such distractors, PolicyLong constructs hard negatives via secondary retrieval. For each verified positive context $B$ (passing Eq.~\ref{eq:entropy_reduction} for position $x_i$), we use it as a query against the database to retrieve top-$k$ semantically similar candidates $C$.

These candidates act as potent hard negatives: they are textually and semantically highly similar to the true supporting evidence $B$, yet they are intrinsically distractors because they are merely nearest-neighbor retrieved texts rather than the actual verified document. Since the positive context $B$ that seeds this secondary retrieval is determined by the current model $M_t$'s entropy landscape, the resulting hard negatives remain fully on-policy and become progressively harder as $M_t$ evolves.

\noindent\textbf{Training sequence construction.}

We construct the final training sequence by assembling the target document $A$, all $m$ verified positive contexts $\{B_1, \ldots, B_m\}$, and their corresponding secondary-retrieved hard negatives $\{C_j^{(1)}, \ldots, C_j^{(k)}\}$ for each positive $B_j$. Crucially, instead of padding with random external texts, all these chunks are \emph{globally shuffled} and prepended to $A$:
\begin{equation}
[\text{globally shuffled } \{B_1, \ldots, B_m, C_1^{(1)}, \ldots, C_m^{(k)}\}]\;A
\label{eq:training_seq}
\end{equation}
By globally shuffling the positive contexts into a massive cluster of highly confusing, secondary-retrieved hard negatives, we force the model to genuinely read and comprehend the semantics to select the correct evidence, rather than relying on superficial similarity or positional recency shortcuts \citep{liu2023lost}.

\textbf{Screening-training distance decoupling.}
We decouple the sequence length used for screening from that used for training. Screening uses shorter contexts (e.g., 2K tokens) for efficiency by only evaluating a small subset of candidates, whereas training scales up to the full target length. To reach the long target length, we rely entirely on these retrieved candidates. We dynamically adjust the secondary retrieval depth: if a document has fewer verified positive candidates, we proportionally increase $k$ (the number of hard negatives retrieved per positive candidate). In this way, the total volume of the globally shuffled positive and hard-negative chunks naturally fills the entire required context window. This decoupling is justified because the contrastive property of hard negatives is preserved---and empirically even amplified---as the overall distance and distractor volume increase.

After constructing the full-length sequence entirely from these shuffled chunks, we train with the standard next-token prediction loss, maintaining simplicity and fair comparability with EntropyLong and NExtLong. Full pseudocode is in Appendix~\ref{app:algorithm}, and Appendix~\ref{app:overhead} discusses the computational efficiency and pipelining of this process, showing that the effective overhead is near zero.

\section{Experiments}
\label{sec:experiments}

\subsection{Experimental setup}

\textbf{Base model.}
We use Qwen2.5-3B as the base model, with an initial context window size of 32K tokens, targeting a 128K context window through long-context extension training.

\textbf{Iterative configuration.}
We train for $T=4$ stages, with 1B tokens per stage, yielding 4B total training tokens. For fair comparison, all baselines use the same total token budget (4B tokens in a single stage).

\textbf{Baselines.}
We compare against the two most recent competitive baselines: (1) NExtLong \citep{nextlong2025}, which uses embedding-based heuristic hard negatives; (2) EntropyLong \citep{jia2025entropylong}, the single-stage entropy-reduction method.

\textbf{Instruction tuning for downstream evaluation.}
We evaluate instruction-following long-context benchmarks only after supervised fine-tuning (SFT). Concretely, after long-context extension training, we further apply the same SFT recipe to all compared models using LongMagpie \citep{gao2025longmagpie}, a self-synthesis method that generates large-scale long-context instruction data, and report LongBench-v2 in this post-SFT setting.

\textbf{Evaluation benchmarks.}
We organize evaluation into two groups based on whether SFT is applied. (1)~\emph{Pre-SFT long-context evaluation}: RULER \citep{hsieh2024ruler} at 8K--128K for synthetic long-range retrieval and reasoning, and HELMET \citep{yen2024helmet} at 8K--128K for holistic long-context understanding. Both are evaluated directly on base models before SFT to measure intrinsic long-context capabilities. (2)~\emph{Post-SFT long-context evaluation}: LongBench-v2 \citep{bai2024longbench2} for instruction-following document-level tasks, evaluated after SFT because its format requires instruction-tuned models (see Appendix~\ref{appendix:benchmarks}).

\subsection{Main results}

\noindent\textbf{Long-context performance beyond 32K.}

Table~\ref{tab:main_long} presents results on RULER (before SFT), HELMET (before SFT), and LongBench-v2 (after SFT) at context lengths beyond 32K, where the off-policy gap is most pronounced.
On RULER, PolicyLong outperforms EntropyLong at both lengths, with the margin widening as context grows: +1.21 at 64K and +2.54 at 128K. This widening gap is consistent with the on-policy hypothesis: longer contexts amplify the misalignment between static training data and the model's evolving capabilities, making on-policy data evolution increasingly valuable.
On HELMET, evaluated on the base models before SFT, PolicyLong similarly leads by +2.63 at 64K and +2.49 at 128K, confirming that the on-policy advantage acquired during continual pre-training directly enhances document-level comprehension.\footnote{Unlike RULER, the on-policy margin on HELMET does not widen with context length. We hypothesize that current pre-training evaluation does not fully elicit the scaling advantage observed in highly structured tasks; exploring more challenging unstructured tasks is left to future work.}
On LongBench-v2, PolicyLong achieves notable gains over EntropyLong on Medium (+2.9) and Long (+2.8)---the subsets requiring handling 32K--2M input contexts. This matches the iterative self-curriculum: as later-stage screening concentrates on the model's true frontier, the training data better prepares it for hard downstream tasks.

\begin{table*}[t]
\caption{Long-context benchmark results ($>$32K) on Qwen2.5-3B. PolicyLong consistently outperforms all baselines, with the advantage widening at longer contexts (e.g., +2.54 over EntropyLong at 128K on RULER). RULER and HELMET are evaluated before SFT; LongBench-v2 is evaluated after SFT. LongBench-v2 Med.: 32K--128K input; Long: 128K--2M input.}
\label{tab:main_long}
\centering
\small
\begin{tabular}{l ccc c ccc c ccc}
\toprule
& \multicolumn{3}{c}{\textbf{RULER}} && \multicolumn{3}{c}{\textbf{HELMET}} && \multicolumn{3}{c}{\textbf{LongBench-v2}} \\
\cmidrule{2-4} \cmidrule{6-8} \cmidrule{10-12}
\textbf{Model} & \textbf{64K} & \textbf{128K} & \textbf{Avg} && \textbf{64K} & \textbf{128K} & \textbf{Avg} && \textbf{Med.} & \textbf{Long} & \textbf{Avg} \\
\midrule
Qwen2.5-3B & 65.36 & 63.22 & 64.29 && 40.53 & 38.99 & 39.76 && 21.5 & 20.6 & 21.1 \\
NExtLong & 70.17 & 61.71 & 65.94 && 45.35 & 41.24 & 43.30 && 23.7 & 26.9 & 25.3 \\
EntropyLong & 69.86 & 63.45 & 66.66 && 43.88 & 40.08 & 41.98 && 24.2 & 28.7 & 26.5 \\
\textbf{PolicyLong} & \textbf{71.07} & \textbf{65.99} & \textbf{68.53} && \textbf{46.51} & \textbf{42.57} & \textbf{44.54} && \textbf{27.1} & \textbf{31.5} & \textbf{29.3} \\
\bottomrule
\end{tabular}
\end{table*}

\begin{table*}[t]
    \caption{Performance at shorter context lengths ($\leq$32K) on Qwen2.5-3B. PolicyLong achieves the best overall short-context performance among all synthesis methods, confirming that on-policy long-context training does not compromise short-context capability. RULER and HELMET are evaluated before SFT; LongBench-v2 Short is evaluated after SFT.}
    \label{tab:short_context}
    \centering
    \small
    \begin{tabular}{l cccc c cccc c c}
    \toprule
    & \multicolumn{4}{c}{\textbf{RULER}} && \multicolumn{4}{c}{\textbf{HELMET}} && \textbf{LB-v2} \\
    \cmidrule{2-5} \cmidrule{7-10} \cmidrule{12-12}
    \textbf{Model} & \textbf{8K} & \textbf{16K} & \textbf{32K} & \textbf{Avg} && \textbf{8K} & \textbf{16K} & \textbf{32K} & \textbf{Avg} && \textbf{Short} \\
    \midrule
    Qwen2.5-3B & 86.43 & 82.24 & 76.82 & 81.83 && 58.30 & 55.85 & 51.36 & 55.17 && 35.8 \\
    NExtLong & 85.82 & 80.73 & 75.90 & 80.82 && 57.52 & 56.00 & 51.19 & 54.90 && 40.6 \\
    EntropyLong & 85.20 & 80.49 & 75.67 & 80.45 && 58.11 & 54.67 & 48.83 & 53.87 && 40.0 \\
    \textbf{PolicyLong} & \textbf{86.05} & \textbf{82.00} & \textbf{78.23} & \textbf{82.09} && \textbf{59.34} & \textbf{56.64} & \textbf{52.57} & \textbf{56.18} && \textbf{40.6} \\
    \bottomrule
    \end{tabular}
    \end{table*}

\noindent\textbf{Performance at shorter context lengths ($\leq$32K).}

Table~\ref{tab:short_context} reports results on RULER and HELMET at context lengths of 8K, 16K, and 32K, as well as LongBench-v2 on the Short ($\leq$32K) subset. Since Qwen2.5-3B already exhibits strong context handling within its native 32K window, the headroom for further improvement at these lengths is inherently limited. Nevertheless, PolicyLong achieves the best overall short-context performance among all synthesis methods, and after SFT it even slightly surpasses the base model, showing that on-policy long-context training preserves short-context ability.

\subsection{Cross-family and cross-scale validation on OLMo3-7B}
\label{sec:olmo}

To verify that PolicyLong is not tied to a single base model, we additionally train it on OLMo3-7B \citep{olmo2025olmo}, covering both a cross-family (Qwen2.5 $\to$ OLMo3) and a cross-scale (3B $\to$ 7B) setting, under a production-scale budget of 50B tokens (one order of magnitude larger than the 4B controlled budget used on Qwen2.5-3B). All compared methods share the same base $+$ LongMagpie SFT trajectory. Besides NExtLong and EntropyLong, we additionally retrain \emph{Longmino}, a baseline that reuses OLMo3's official long-context stage-extension data under the same pipeline, with long-context ($>$32K) results reported in Table~\ref{tab:olmo}.

\begin{table*}[t]
\caption{Cross-family and cross-scale validation on OLMo3-7B ($>$32K), trained under a production-scale 50B-token budget. PolicyLong achieves the best long-context average on RULER, HELMET, and LongBench-v2, with its strongest gains at the longest context settings. RULER and HELMET are evaluated before SFT; LongBench-v2 is evaluated after SFT. Avg denotes the long-context average: (64K$+$128K)/2 for RULER/HELMET and (Med.$+$Long)/2 for LongBench-v2. $^{\dagger}$Longmino reuses OLMo3's official long-context stage-extension data, retrained under the same pipeline.}
\label{tab:olmo}
\centering
\small
\begin{tabular}{l ccc c ccc c ccc}
\toprule
& \multicolumn{3}{c}{\textbf{RULER}} && \multicolumn{3}{c}{\textbf{HELMET}} && \multicolumn{3}{c}{\textbf{LongBench-v2}} \\
\cmidrule{2-4} \cmidrule{6-8} \cmidrule{10-12}
\textbf{Model} & \textbf{64K} & \textbf{128K} & \textbf{Avg} && \textbf{64K} & \textbf{128K} & \textbf{Avg} && \textbf{Med.} & \textbf{Long} & \textbf{Avg} \\
\midrule
Longmino$^{\dagger}$ & 71.4 & 65.4 & 68.40 && 54.6 & 48.1 & 51.35 && 25.27 & 24.07 & 24.67 \\
NExtLong & \textbf{73.7} & 69.5 & 71.60 && 54.9 & 51.0 & 52.95 && 25.74 & 23.77 & 24.76 \\
EntropyLong & 71.4 & 68.7 & 70.05 && 56.0 & 52.2 & 54.10 && 28.68 & 25.00 & 26.84 \\
\textbf{PolicyLong} & 73.3 & \textbf{72.7} & \textbf{73.00} && \textbf{57.1} & \textbf{54.4} & \textbf{55.75} && \textbf{29.46} & \textbf{29.32} & \textbf{29.39} \\
\bottomrule
\end{tabular}
\end{table*}

On OLMo3-7B, PolicyLong achieves the best long-context average on all three benchmarks, while trailing NExtLong slightly at RULER 64K (73.3 vs.\ 73.7). On RULER it reaches a long-context average of 73.00 (+2.95 over EntropyLong), with the largest gap at 128K (+4.0 over EntropyLong, +3.2 over NExtLong); on HELMET it leads with 55.75 (+1.65 over EntropyLong, +2.80 over NExtLong); and after SFT on LongBench-v2 it attains the best long-context average of 29.39, where the Long (128K--2M) subset improves by +4.32 over EntropyLong and +5.55 over NExtLong. Consistent with the Qwen2.5-3B results, the margin again concentrates at the longest contexts, confirming that the benefit of on-policy data evolution does not depend on a single base model.

\section{Ablations and Analysis}
\label{sec:ablation}

\subsection{Component Ablation}

We conduct systematic ablations to isolate the contribution of each on-policy component (Tables~\ref{tab:ablation_long} and~\ref{tab:ablation_short}). We study three settings: \textbf{Iteration} (whether entropy is re-computed with the updated model $M_t$ at each stage), \textbf{Hard Negatives} (whether hard negatives are included, and if so, whether they are \emph{static}---constructed once with $M_0$---or \emph{on-policy}---reconstructed at each stage with $M_t$), and \textbf{Curriculum} (whether the difficulty threshold adapts dynamically across stages). Unless otherwise specified, all ablations are performed on Qwen2.5-3B.

\begin{table}[t]
\caption{Ablation on RULER at long contexts ($>$32K) for Qwen2.5-3B. Iter.\ = Iteration; EL = EntropyLong. $\checkmark$/$\times$ denotes enabled/disabled.}
\label{tab:ablation_long}
\centering
\small
\begin{tabular}{lcccccc}
\toprule
\textbf{Configuration} & \textbf{Iter.} & \textbf{HN} & \textbf{Curr.} & \textbf{64K} & \textbf{128K} & \textbf{Avg} \\
\midrule
EntropyLong & $\times$ & $\times$ & $\times$ & 69.86 & 63.45 & 66.66 \\
\textbullet~Iteration & $\checkmark$ & $\times$ & $\times$ & 70.66 & 63.62 & 67.14 \\
\textbullet~Hard Negatives & $\times$ & $\checkmark$ & $\times$ & 70.27 & 64.66 & 67.46 \\
\textbullet~Curriculum & $\times$ & $\times$ & $\checkmark$ & 70.01 & 63.75 & 66.88 \\
\textbullet~Iter. + Curriculum & $\checkmark$ & $\times$ & $\checkmark$ & 70.13 & 64.11 & 67.12 \\
\textbullet~Iter. + HN & $\checkmark$ & $\checkmark$ & $\times$ & 69.95 & 65.16 & 67.56 \\
\textbullet~HN + Curriculum & $\times$ & $\checkmark$ & $\checkmark$ & 70.53 & 65.28 & 67.91 \\
\textbf{PolicyLong (Full)} & $\checkmark$ & $\checkmark$ & $\checkmark$ & \textbf{71.07} & \textbf{65.99} & \textbf{68.53} \\
\bottomrule
\end{tabular}
\end{table}

\begin{table}[t]
\caption{Ablation on RULER at short contexts ($\leq$32K) for Qwen2.5-3B. Iter.\ = Iteration; EL = EntropyLong. $\checkmark$/$\times$ denotes enabled/disabled.}
\label{tab:ablation_short}
\centering
\small
\begin{tabular}{lccccccc}
\toprule
\textbf{Configuration} & \textbf{Iter.} & \textbf{HN} & \textbf{Curr.} & \textbf{8K} & \textbf{16K} & \textbf{32K} & \textbf{Avg} \\
\midrule
EntropyLong & $\times$ & $\times$ & $\times$ & 85.20 & 80.49 & 75.67 & 80.45 \\
\textbullet~Iteration & $\checkmark$ & $\times$ & $\times$ & 85.74 & 80.65 & 75.92 & 80.77 \\
\textbullet~Hard Negatives & $\times$ & $\checkmark$ & $\times$ & 85.30 & 81.44 & 76.49 & 81.08 \\
\textbullet~Curriculum & $\times$ & $\times$ & $\checkmark$ & 85.13 & 81.04 & 75.71 & 80.63 \\
\textbullet~Iter. + Curriculum & $\checkmark$ & $\times$ & $\checkmark$ & 85.61 & 81.23 & 76.51 & 81.12 \\
\textbullet~Iter. + HN & $\checkmark$ & $\checkmark$ & $\times$ & 85.81 & 81.68 & 76.51 & 81.33 \\
\textbullet~HN + Curriculum & $\times$ & $\checkmark$ & $\checkmark$ & 85.24 & 81.68 & 77.19 & 81.37 \\
\textbf{PolicyLong (Full)} & $\checkmark$ & $\checkmark$ & $\checkmark$ & \textbf{86.05} & \textbf{82.00} & \textbf{78.23} & \textbf{82.09} \\
\bottomrule
\end{tabular}
\end{table}

\textbf{Effect of iterative model updates.}
Enabling \emph{Iteration} alone raises the $>$32K average from 66.66 to 67.14, with a modest $\leq$32K gain (80.45 $\to$ 80.77). Re-computing entropy with the updated model refreshes the positive training signal, but without an increasing threshold or hard negatives the re-screened data still spans a similar difficulty distribution, limiting the gain. \emph{Curriculum} alone barely moves the $>$32K average (66.88), since without iterative re-screening its threshold schedule operates on the fixed base-model entropy landscape; only when combined with \emph{Iteration} does it take effect, lifting the $\leq$32K average to 81.12 (67.12 at $>$32K). The stage-increasing threshold $\tau$ then ensures that each stage focuses on the updated model's true high-entropy frontier rather than recycling positions already mastered, allowing the self-curriculum to take effect.

\textbf{Effect of hard negatives.}
\emph{Hard Negatives} are the strongest single component: enabling HN alone already lifts the $>$32K average to 67.46 ($+0.80$), confirming that the contrastive signal---training the model to resist plausible distractors---provides a stronger optimization benefit than refining positive-context selection alone. Adding \emph{On-policy HN} on top of \emph{Iteration} raises the $>$32K average to 67.56. Notably, the \emph{HN + Curriculum} configuration (static hard negatives screened once by $M_0$, without \emph{Iteration}) reaches 67.91, the strongest result among the partial configurations, showing that introducing hard negatives is beneficial even when static. However, only when we enable \emph{Iteration} does this contrastive signal become truly \emph{on-policy}, co-evolving with the current model to yield the largest overall improvement.

\textbf{Complementarity of components.}
The \emph{full PolicyLong} configuration achieves the best results on both groups: 68.53 ($>$32K) and 82.09 ($\leq$32K). The three components address orthogonal axes of the on-policy principle---\emph{Iteration} aligns the positive training signal, \emph{Curriculum} adapts the difficulty distribution, and \emph{On-policy HN} co-evolves the contrastive signal---and their combination yields the largest improvement.

\begin{figure}[t]
    \centering
    \begin{subfigure}[t]{0.48\textwidth}
        \centering
        \includegraphics[width=0.8\linewidth]{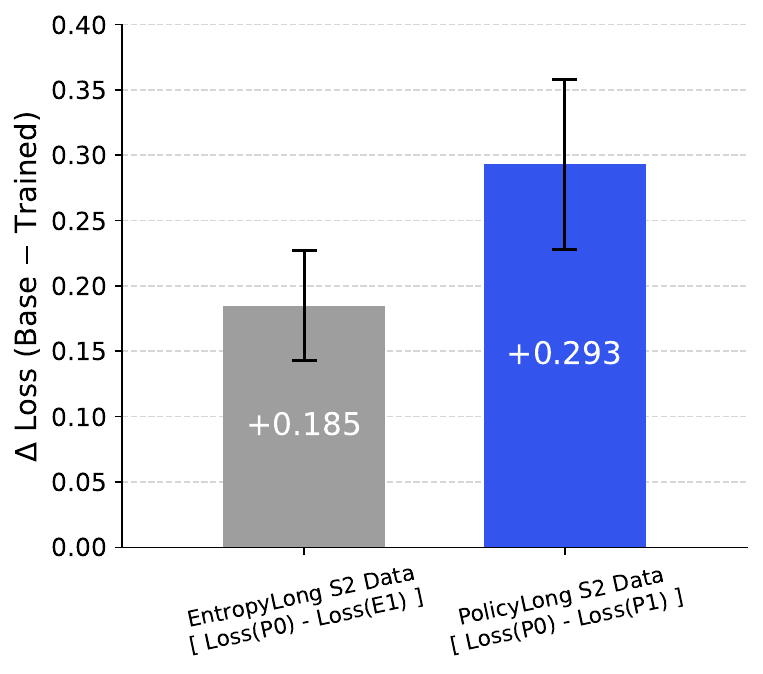}
        \caption{Stage 2 data difficulty.}
        \label{fig:data_difficulty_s2}
    \end{subfigure}
    \hfill
    \begin{subfigure}[t]{0.48\textwidth}
        \centering
        \includegraphics[width=0.8\linewidth]{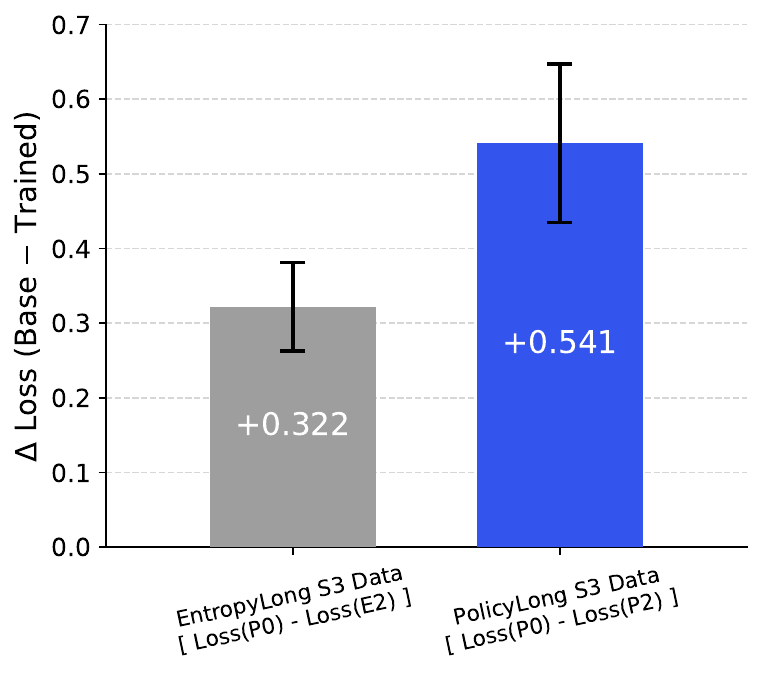}
        \caption{Stage 3 data difficulty.}
        \label{fig:data_difficulty_s3}
    \end{subfigure}
    \caption{Data difficulty progression. We measure the loss reduction from the base model to the progressively trained model on data generated at Stage 2 and Stage 3. PolicyLong yields a significantly larger loss gap than EntropyLong, indicating that it successfully surfaces harder, non-trivial dependencies.}
    \label{fig:data_difficulty}
\end{figure}

\subsection{Difficulty Progression of Constructed Data}

To verify that the on-policy mechanism surfaces progressively harder dependencies, we define a \emph{data difficulty} metric: for data generated at stage $t$, we compute the loss reduction $\text{Loss}(P_0) - \text{Loss}(M_{t-1})$, where $P_0$ is the base model and $M_{t-1}$ is the model trained up to the previous stage ($P_{t-1}$ for PolicyLong, $E_{t-1}$ for EntropyLong). A larger value indicates that the constructed dependencies are harder for the base model yet learnable by the trained model---exactly the property that effective on-policy data should exhibit. As shown in Figure~\ref{fig:data_difficulty}, at Stage~2 PolicyLong yields $\text{Loss}(P_0) - \text{Loss}(P_1) = +0.293$, versus EntropyLong's $\text{Loss}(P_0) - \text{Loss}(E_1) = +0.185$; at Stage~3 the margin widens to $+0.541$ versus $+0.322$. This compounding effect confirms that the on-policy closed loop pushes the training distribution toward increasingly challenging dependencies that static screening fails to capture.

\subsection{General Short-Text Capability Retention}

Table~\ref{tab:short_text} compares the base model and the full PolicyLong model on general short-text benchmarks for Qwen2.5-3B. The purpose of this comparison is not to claim gains on short-text tasks, but to verify that the proposed long-context training recipe does not erode general short-context reasoning and commonsense ability. Consistent with EntropyLong, PolicyLong preserves essentially the same average short-text performance (59.04 vs.\ 58.99), without sacrificing general short-context ability.

\begin{table}[t]
\caption{General short-text benchmark results on Qwen2.5-3B. We compare the base model and the full PolicyLong model to verify that long-context extension does not degrade general short-text capabilities.}
\label{tab:short_text}
\centering
\small
\begin{tabular}{lccccccc}
\toprule
\textbf{Model} & \textbf{ARC-C} & \textbf{ARC-E} & \textbf{HellaSwag} & \textbf{PIQA} & \textbf{LogiQA} & \textbf{WinoGrande} & \textbf{Avg} \\
\midrule
Qwen2.5-3B & 45.04 & 77.36 & 55.00 & 78.24 & 29.95 & 68.35 & 58.99 \\
\textbf{PolicyLong} & 45.90 & 78.45 & 53.97 & 78.02 & 29.95 & 67.96 & \textbf{59.04} \\
\bottomrule
\end{tabular}
\end{table}

\raggedbottom
\section{Related work}
\label{sec:related}

\textbf{Long-context data construction.}
Since naturally long documents are rare in web-scale corpora, recent work synthesizes long-context data by concatenating short documents. Early heuristics either randomly concatenate documents \citep{xiong2023effective} or sort them by embedding similarity \citep{shi2024incontext}, but lack verification of genuine dependencies. More targeted methods---query-centric synthesis (Quest, \citealp{gao2024quest}), dependency-recognition synthesis (Re3Syn, \citealp{zhang2025re3syn}), and resource-efficient synthesis (LiteLong, \citealp{jia2026litelong})---still rely on heuristic verification. EntropyLong \citep{jia2025entropylong} instead adopts entropy-reduction verification, checking whether a candidate context significantly reduces the base model's predictive entropy at high-uncertainty positions. Yet all these methods build the dataset in a single offline pass with a fixed model, so the static distribution grows misaligned with the evolving model---the off-policy gap. PolicyLong bridges it by turning data construction into an iterative, on-policy closed loop.

\textbf{On-policy learning and implicit self-curriculum.}
On-policy learning holds that models learn more effectively from feedback derived from their current policy than from fixed, off-policy data. This has been validated across diverse settings, including on-policy self-distillation for reasoning \citep{zhao2026self}, continual learning \citep{shenfeld2026self}, credit assignment in RL \citep{hubotter2026reinforcement}, and privileged exploration for hard reasoning \citep{qu2026pope}. Such learning naturally induces an implicit curriculum---as feedback evolves with the model, training difficulty progresses accordingly---echoing curriculum \citep{bengio2009curriculum} and active learning \citep{settles2009active}. Yet this principle remains unexplored in long-context data construction, which relies entirely on a fixed base model. PolicyLong brings it to this domain: the model's own entropy landscape serves as the on-policy signal, and the resulting self-curriculum needs no explicit scheduling---early stages surface basic dependencies while later stages focus on the model's learning frontier.

\textbf{Hard negative mining.}
Hard negative mining is well-established in contrastive learning \citep{chen2020simclr,robinson2020hard} and dense retrieval, where strategies have evolved from static negatives \citep{karpukhin2020dense} to dynamic ones that refresh with the model \citep{xiong2020ance,qu2021rocketqa}. In long-context training, NExtLong \citep{nextlong2025} mines embedding-similar distractors as hard negatives, but selects them once with a fixed model---a static, off-policy criterion. PolicyLong instead constructs hard negatives via secondary retrieval seeded by the current model $M_t$'s verified positives. Since these positives are determined by $M_t$'s entropy landscape, the negatives co-evolve with the model, forming a unified on-policy mechanism governing both what the model exploits and what it resists.

\flushbottom
\section{Conclusion}

PolicyLong reformulates long-context data construction as an iterative on-policy process rather than a fixed offline step. By re-screening data with the current model, it keeps both positive contexts and hard negatives aligned with the model's evolving frontier and induces an implicit self-curriculum across stages. Experiments on Qwen2.5-3B and OLMo3-7B show consistent gains over static baselines on long-context benchmarks while preserving short-text capability, with the advantage holding across model family and scale. Overall, the results suggest that on-policy data construction is a more effective paradigm for long-context extension than one-shot offline screening.

\section{Limitations}

Our study validates PolicyLong on Qwen2.5-3B and OLMo3-7B, spanning two model families and two scales (3B and 7B) as well as both controlled (4B-token) and production-scale (50B-token) budgets. Extending the framework to substantially larger models ($\geq$30B) and to even longer target windows ($\geq$256K) remains computationally demanding and is left to future work. In addition, while the iterative refresh delivers consistent gains, the marginal benefit diminishes at larger stage counts (e.g., $T{=}5$); designing adaptive refresh schedules that keep improving in this regime is a promising direction.

\bibliography{example_paper}
\bibliographystyle{iclr2025_conference}

\appendix
\section{Additional experimental details}
\label{app:details}

\textbf{Retrieval corpus.}
We use FineWeb-Edu \citep{lozhkov2024fineweb-edu} and Cosmopedia \citep{benallal2024cosmopedia} as retrieval corpora, following the setup of EntropyLong.

\textbf{Retrieval model.}
We use Jina Embeddings v3 \citep{sturua2024jina} for dense retrieval and FAISS \citep{johnson2019billion} for efficient nearest-neighbor search.

\textbf{SFT protocol for downstream benchmarks.}
We use LongMagpie \citep{gao2025longmagpie} for supervised fine-tuning before evaluating LongBench-v2. LongMagpie is a self-synthesis method that automatically generates large-scale long-context instruction data without human annotation. RULER and HELMET are evaluated before SFT, while LongBench-v2 is evaluated after SFT under the same recipe for all compared methods.

\section{PolicyLong algorithm}
\label{app:algorithm}

Algorithm~\ref{alg:policylong} presents the complete pseudocode for PolicyLong. The outer loop iterates over $T$ training stages. At each stage $t$, a fresh set of root documents $\mathcal{D}_t$ is sampled, and the current model $M_t$ is used to execute the full data construction pipeline: computing per-token entropy, identifying high-entropy positions via the stage-increasing threshold $\tau_t$, retrieving candidate contexts, and verifying positive contexts through entropy-reduction (Eq.~\ref{eq:entropy_reduction}). For each verified positive $B$, a secondary retrieval step produces hard negatives by querying the corpus with $B$ itself, and the resulting positive and hard-negative chunks are globally shuffled and prepended to the root document $A$ to form a full-length training sequence. The secondary retrieval depth is dynamically adjusted so that the total volume of shuffled chunks fills the target sequence length $L$. After collecting $N$ tokens of training data $\mathcal{S}_t$, the model is updated to $M_{t+1}$ via standard next-token prediction training, and the process repeats with the updated model.

\begin{algorithm}[t]
    \caption{PolicyLong: On-Policy Long-Context Training}
    \label{alg:policylong}
    \begin{algorithmic}[1]
    \REQUIRE Source corpus $\mathcal{D}$, Retrieval corpus $\mathcal{R}$, Base model $M_0$, stages $T$, tokens per stage $N$, sequence length $L$, entropy thresholds $\{\tau_t\}_{t=0}^{T-1}$ increasing with $t$
    \ENSURE Long-context model $M_T$
    \FOR{$t = 0$ to $T-1$}
        \STATE Sample document set $\mathcal{D}_t$ from $\mathcal{D}$
        \FOR{root document $A \in \mathcal{D}_t$}
            \STATE Compute per-token entropy using $M_t$
            \STATE Set the stage entropy threshold $\tau_t \leftarrow 0.2(t+1)$, increasing with $t$ ($\tau_t \in \{0.2,0.4,0.6,0.8\}$ for $T=4$)
            \STATE Select positions whose entropy exceeds $\tau_t$
            \STATE Retrieve $K$ candidate contexts from $\mathcal{R}$ using fragments around those positions
            \STATE Verify the positive-context set $\mathcal{B}=\{B_1,\ldots,B_m\}$ satisfying $\Delta H(B_j, x_i) > \tau_{\text{pos}}$
            \STATE Initialize the hard-negative collection $\mathcal{C} \leftarrow \varnothing$
            \STATE Set secondary retrieval depths $\{k_j\}_{j=1}^{m}$ so that the combined chunks fill target length $L$
            \FOR{each verified positive context $B_j \in \mathcal{B}$}
                \STATE Retrieve $\mathcal{C}_j=\{C_j^{(1)},\ldots,C_j^{(k_j)}\}$ using $B_j$ as the query
                \STATE Update $\mathcal{C} \leftarrow \mathcal{C} \cup \mathcal{C}_j$
            \ENDFOR
            \STATE Construct one full-length sequence $[\text{globally shuffled }(\mathcal{B}\cup\mathcal{C})]\;A$
        \ENDFOR
        \STATE Collect training data $\mathcal{S}_t$ ($N$ tokens)
        \STATE Train: $M_{t+1} \leftarrow \text{Train}(M_t, \mathcal{S}_t)$
    \ENDFOR
    \RETURN $M_T$
    \end{algorithmic}
\end{algorithm}

\section{Full benchmark results}
\label{app:full_results}

\subsection{RULER results across target context lengths}

Table~\ref{tab:ruler_target_lengths} compares PolicyLong and EntropyLong on RULER when trained with different target context lengths (64K, 128K, and 256K). At every target length, PolicyLong consistently outperforms the corresponding EntropyLong baseline, with the margin most pronounced at longer evaluation lengths. Notably, scaling the training target from 128K to 256K further improves PolicyLong's performance, achieving the highest overall average (77.24), confirming that the on-policy mechanism remains effective across different training configurations.

\begin{table*}[t]
\caption{RULER benchmark results on Qwen2.5-3B across different training target context lengths. All results are at checkpoint-500.}
\label{tab:ruler_target_lengths}
\centering
\small
\begin{tabular}{llcccccc}
\toprule
\textbf{Target Length} & \textbf{Model} & \textbf{8K} & \textbf{16K} & \textbf{32K} & \textbf{64K} & \textbf{128K} & \textbf{Avg} \\
\midrule
\multirow{2}{*}{64K}
& EntropyLong & 83.71 & 79.57 & 74.56 & 68.37 & 59.69 & 73.18 \\
& \textbf{PolicyLong} & \textbf{85.26} & \textbf{80.82} & \textbf{76.58} & \textbf{70.14} & \textbf{61.66} & \textbf{74.89} \\
\midrule
\multirow{2}{*}{128K}
& EntropyLong & 85.20 & 80.49 & 75.67 & 69.86 & 63.45 & 74.93 \\
& \textbf{PolicyLong} & \textbf{86.05} & \textbf{82.00} & \textbf{78.23} & \textbf{71.07} & \textbf{65.99} & \textbf{76.67} \\
\midrule
\multirow{2}{*}{256K}
& EntropyLong & 85.81 & 81.48 & 76.95 & 71.66 & 65.87 & 76.35 \\
& \textbf{PolicyLong} & \textbf{86.24} & \textbf{82.24} & \textbf{78.34} & \textbf{72.56} & \textbf{66.81} & \textbf{77.24} \\
\bottomrule
\end{tabular}
\end{table*}

\subsection{Needle-in-a-Haystack evaluation}

\begin{figure*}[t]
    \centering
    \includegraphics[width=\textwidth]{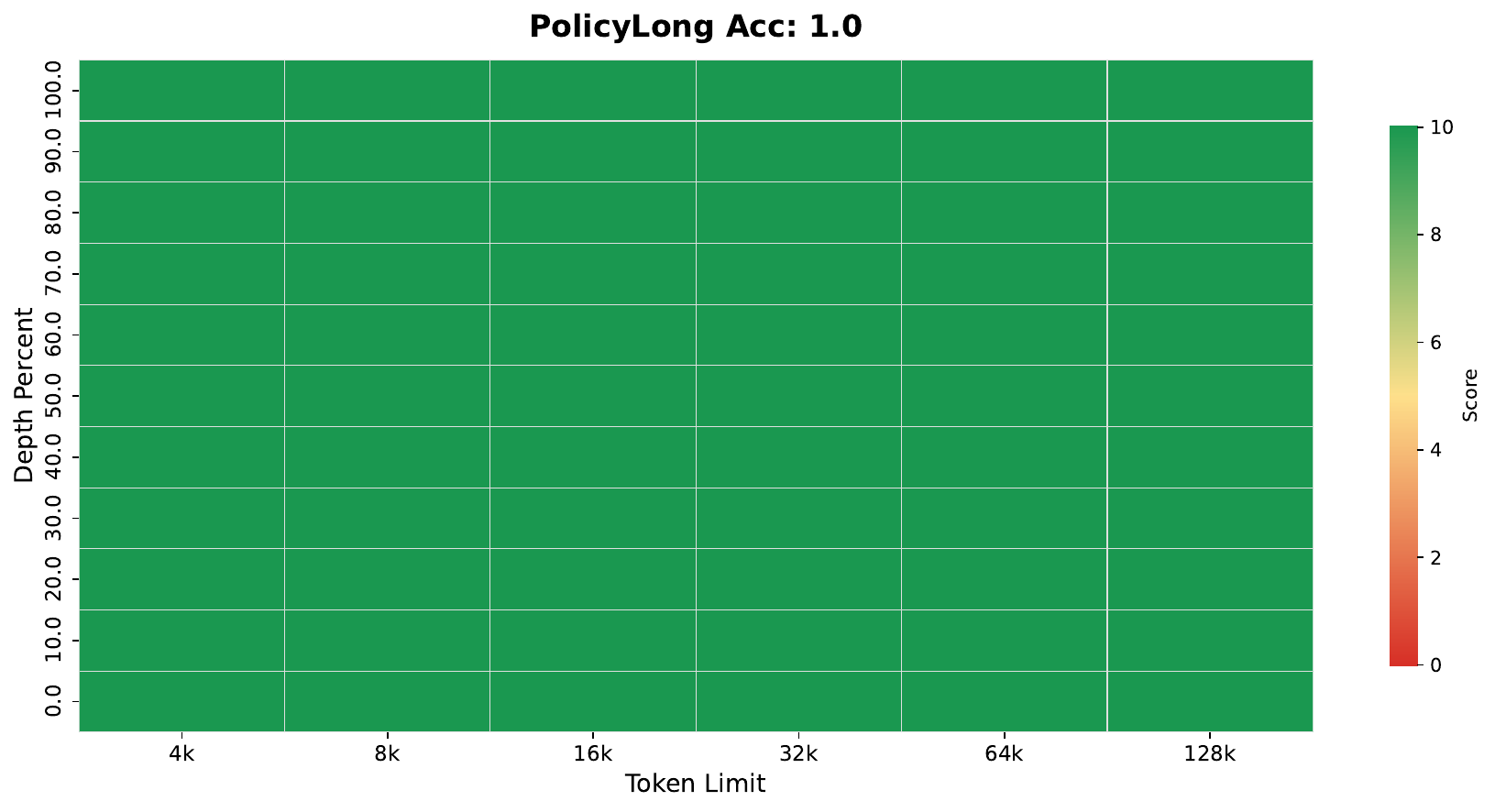}
    \caption{Needle-in-a-Haystack evaluation results for PolicyLong. The heatmap shows retrieval accuracy across varying context lengths (x-axis) and needle depth positions (y-axis).}
    \label{fig:niah_heatmap}
\end{figure*}

Figure~\ref{fig:niah_heatmap} shows the Needle-in-a-Haystack (NIAH) evaluation results for PolicyLong. The heatmap visualizes retrieval accuracy across different context lengths and needle positions, demonstrating that PolicyLong maintains strong retrieval performance across the full range of context lengths and depth positions.

\section{Benchmark Details}
\label{appendix:benchmarks}

\textbf{RULER Benchmark Tasks:}
\begin{itemize}[itemsep=0.1em,parsep=0pt,topsep=0.3em]
\item \textbf{Needle-in-a-Haystack (NIAH)}: A family of four retrieval-oriented tasks, including Single NIAH (standard single-needle retrieval), Multi-keys NIAH (finding the correct needle among several distractor keys), Multi-values NIAH (recovering all values associated with one key), and Multi-queries NIAH (retrieving multiple different needles)
\item \textbf{Variable Tracking (VT)}: A multi-hop tracing task for coreference reasoning, where the model follows chains of variable assignments (e.g., X2 = X1, X3 = X2) and identifies all variables linked to the same original value
\item \textbf{Common Words Extraction (CWE) and Frequent Words Extraction (FWE)}: Aggregation-style tasks in which the model must determine the most frequent words from long sequences, acting as proxies for summarization when relevant evidence is distributed across broad context spans
\item \textbf{Question Answering (QA)}: A realistic variant of NIAH in which answer-containing paragraphs are embedded among distractor paragraphs drawn from the same dataset, and the question serves as the retrieval query
\end{itemize}

\textbf{LongBench-v2 Tasks:}
\begin{itemize}[itemsep=0.1em,parsep=0pt,topsep=0.3em]
\item \textbf{Single-Document QA}: Question answering over individual long documents from six domains (Academic, Literary, Legal, Financial, Governmental, Detective), together with event ordering tasks, evaluating document-level understanding
\item \textbf{Multi-Document QA}: Reasoning across multiple documents, requiring the model to combine evidence from different sources in domains such as Academic, Legal, Financial, Governmental, and Multi-news
\item \textbf{Long In-context Learning}: A set of three difficult tasks, including user guide QA, translation for new languages (Zhuang and Kalamang), and many-shot classification with anonymized labels
\item \textbf{Long-dialogue History Understanding}: Understanding long conversational histories, including agent interaction games (GAMA-Bench) and multi-turn dialogue sessions (LongMemEval), where success depends on retaining distant context
\item \textbf{Code Repository Understanding}: Comprehensive understanding of large code repositories, requiring the model to connect multiple code components and reason about complex software structure
\item \textbf{Long Structured Data Understanding}: Reasoning over large structured inputs such as financial tables and large knowledge graphs, with an emphasis on multi-hop inference over interconnected entities
\end{itemize}

\section{Computational overhead}
\label{app:overhead}

Compared with EntropyLong, PolicyLong's only additional cost is the \emph{re-execution of the data construction pipeline} at stages $t \geq 1$. This pipeline consists of three steps: (1)~computing per-token entropy with $M_t$ to identify high-entropy positions, (2)~retrieving candidate contexts from a pre-built FAISS index, and (3)~verifying entropy reduction. Step~(2), which dominates the pipeline latency, is entirely a \textbf{CPU-bound} operation: it involves encoding query fragments with a frozen embedding model and performing approximate nearest-neighbor search over a static FAISS index---no GPU is required. Steps~(1) and~(3) require forward passes through $M_t$, but these are inference-only operations on short sequences that are far cheaper than training-time backward passes on long sequences.

Crucially, the data construction for stage~$t{+}1$ can be \textbf{pipelined with} the GPU training of stage~$t$: while the GPU is occupied with gradient updates on $\mathcal{S}_t$, the CPU retrieval and lightweight GPU verification for $\mathcal{S}_{t+1}$ proceed in parallel. In our setup with $T{=}4$ stages, each stage's data construction completes well within the training wall-clock time of the preceding stage, so the effective overhead is near zero. Overall, PolicyLong's total GPU training cost (tokens $\times$ FLOPs) is identical to EntropyLong's, with the iterative re-screening adding only a marginal CPU cost that is fully hidden by pipelining.

\section{RULER subtask analysis}
\label{app:ruler_analysis}

\begin{figure}[t]
    \centering
    \begin{subfigure}[t]{0.48\textwidth}
        \centering
        \includegraphics[width=\linewidth]{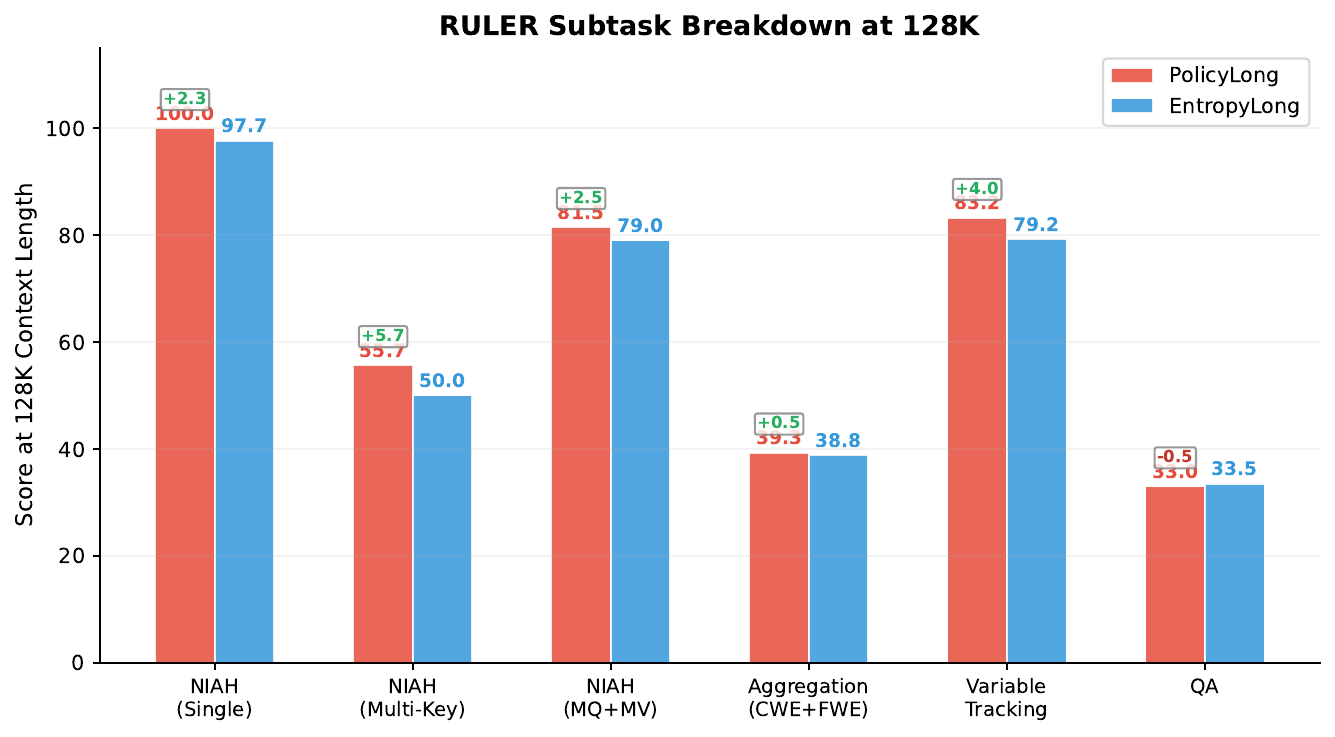}
        \caption{RULER subtask breakdown at 128K.}
        \label{fig:ruler_128k}
    \end{subfigure}
    \hfill
    \begin{subfigure}[t]{0.48\textwidth}
        \centering
        \includegraphics[width=\linewidth]{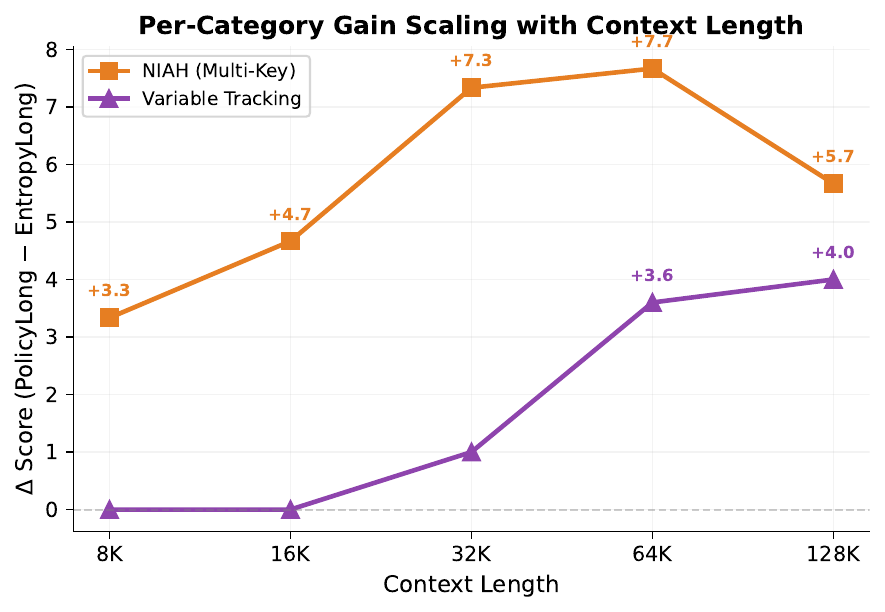}
        \caption{Per-category gain scaling with context length.}
        \label{fig:ruler_gain_scaling}
    \end{subfigure}
    \caption{RULER detailed analysis. (a)~Subtask breakdown at 128K: PolicyLong achieves the largest gains on NIAH Multi-Key (+5.7) and Variable Tracking (+4.0). (b)~Gain (PolicyLong $-$ EntropyLong) scaling with context length on the two most improved subtasks.}
    \label{fig:ruler_detailed}
\end{figure}

Figure~\ref{fig:ruler_detailed}(a) presents a per-subtask comparison between PolicyLong and EntropyLong on RULER at 128K. The gains are concentrated on the tasks that require the most precise long-range reasoning: \textbf{NIAH Multi-Key} (+5.7) demands distinguishing the correct key--value pair from multiple similar distractors scattered across the context, and \textbf{Variable Tracking} (+4.0) requires multi-hop coreference resolution over long variable assignment chains. Both tasks are particularly sensitive to the quality of long-range dependencies in the training data. PolicyLong's on-policy hard negatives---semantically similar chunks that force the model to attend to fine-grained distinctions---directly train the capability exercised by these subtasks. In contrast, \textbf{Aggregation} (+0.5) and \textbf{QA} ($-$0.5) show marginal differences, as these subtasks rely more on holistic summarization or localized comprehension where the off-policy gap is less pronounced.

Figure~\ref{fig:ruler_detailed}(b) further examines how the PolicyLong advantage scales with context length on the two most improved subtasks. For \textbf{NIAH Multi-Key}, the gain increases steadily from +3.3 at 8K to +7.7 at 64K before settling at +5.7 at 128K, confirming that the on-policy mechanism becomes increasingly valuable as the context grows and the search space for the correct key--value pair expands. \textbf{Variable Tracking} exhibits a similar upward trend. The consistent widening of gains at longer contexts aligns with our central hypothesis: static, off-policy data fails to capture the evolving difficulty landscape at scale, whereas PolicyLong's iterative re-screening continuously adapts the training distribution to the model's frontier.

\section{Statistical Significance and Seed Variance}
\label{app:significance}

To verify that the improvement of PolicyLong over EntropyLong is statistically reliable rather than an artifact of a single training run, we retrain both methods with three independent training seeds (42, 43, 44) on Qwen2.5-3B and report the RULER 8K--128K average in Table~\ref{tab:seed_variance}. RULER and HELMET are evaluated with temperature $0$, so repeated evaluations of the same checkpoint are deterministic; the variance reported here therefore reflects \emph{training-seed} variance, which is the relevant source of randomness for our claims.

\begin{table}[t]
\caption{Multi-seed RULER average (8K--128K) on Qwen2.5-3B. Each method is trained with three independent seeds. The per-method standard deviation ($\leq$0.16) is far smaller than the mean improvement of PolicyLong over EntropyLong (+1.67).}
\label{tab:seed_variance}
\centering
\small
\begin{tabular}{lccccc}
\toprule
\textbf{Method} & \textbf{seed 42} & \textbf{seed 43} & \textbf{seed 44} & \textbf{mean} & \textbf{std} \\
\midrule
EntropyLong & 74.93 & 75.15 & 74.84 & 74.97 & 0.16 \\
\textbf{PolicyLong} & 76.67 & 76.60 & 76.65 & \textbf{76.64} & \textbf{0.04} \\
\midrule
$\Delta$ (PolicyLong $-$ EntropyLong) & +1.74 & +1.45 & +1.81 & +1.67 & -- \\
\bottomrule
\end{tabular}
\end{table}

We perform a \emph{paired} $t$-test over the three seed-matched pairs (the two methods sharing the same training seed). The mean improvement is $+1.67$, with paired $t(2)=15.1$ and two-tailed $p\approx0.004$ ($<0.05$), and a $95\%$ confidence interval of $[1.19,\,2.14]$ that excludes $0$. The improvement is thus significant at the $95\%$ level, and the mean gap far exceeds each method's within-group standard deviation ($\leq 0.16$), confirming that the advantage of on-policy data evolution is not attributable to seed noise.

\section{Effect of the Number of Stages \texorpdfstring{$T$}{T}}
\label{app:num_stages}

The number of on-policy refresh stages $T$ is the central hyperparameter of PolicyLong ($T=4$ in the main configuration). To characterize its effect, we sweep $T \in \{1,2,3,4,5\}$ while holding the total training token budget fixed, so that larger $T$ corresponds to more frequent (but smaller) refreshes rather than more total training. $T=1$ recovers the off-policy EntropyLong baseline. Table~\ref{tab:num_stages} reports the RULER 8K--128K average on Qwen2.5-3B.

\begin{table}[t]
\caption{Effect of the number of on-policy refresh stages $T$ (RULER 8K--128K average, Qwen2.5-3B), at a fixed total token budget. $T=1$ corresponds to the off-policy EntropyLong baseline.}
\label{tab:num_stages}
\centering
\small
\begin{tabular}{lcc}
\toprule
\textbf{$T$ (stages)} & \textbf{RULER Avg} & \textbf{$\Delta$ vs $T=1$} \\
\midrule
1 (EntropyLong) & 74.93 & -- \\
2 & 76.11 & +1.18 \\
3 & 76.55 & +1.62 \\
4 (main config) & \textbf{76.67} & \textbf{+1.74} \\
5 & 76.45 & +1.52 \\
\bottomrule
\end{tabular}
\end{table}

A single additional on-policy refresh stage ($T{=}2$) already yields a large $+1.18$ gain, and adding more stages continues to help up to $T{=}4$. Beyond that the benefit saturates: $T{=}5$ no longer improves over $T{=}4$. We attribute this to diminishing returns once the refresh frequency is high enough to keep the training distribution closely aligned with the evolving model, at which point the per-stage data volume (and thus the optimization signal per refresh) becomes the limiting factor under a fixed budget. Designing adaptive refresh schedules (e.g., a varying refresh frequency or threshold) that keep improving at larger $T$ is an interesting direction for future work.

\section{Hard Negative Design Choices}
\label{app:hn_design}

PolicyLong constructs hard negatives via secondary semantic-similarity retrieval seeded by the on-policy verified positives. Here we isolate two questions: (i)~does the \emph{difficulty} (semantic similarity) of the hard negatives matter, and (ii)~is the gain explained simply by negative \emph{diversity} or by a particular mining algorithm? All variants below keep the rest of the PolicyLong pipeline fixed and change only the hard-negative source; we report the RULER 8K--128K average on Qwen2.5-3B (Table~\ref{tab:hn_design}).

\begin{table}[t]
\caption{Hard-negative design choices on Qwen2.5-3B (RULER 8K--128K average). Only the hard-negative source is varied; all other components match the full PolicyLong pipeline. The BM25 pair differs only in fixed (off-policy) vs.\ on-policy refresh.}
\label{tab:hn_design}
\centering
\small
\begin{tabular}{lcc}
\toprule
\textbf{Hard-negative source} & \textbf{RULER Avg} & \textbf{$\Delta$ vs PolicyLong} \\
\midrule
Random chunks (no secondary retrieval) & 73.77 & $-2.90$ \\
Per-query random (higher diversity) & 74.20 & $-2.47$ \\
Bottom-$k$ retrieval (least similar) & 75.48 & $-1.19$ \\
\midrule
BM25, fixed (off-policy) & 74.01 & $-2.66$ \\
BM25, on-policy refresh & 74.46 & $-2.21$ \\
\midrule
\textbf{PolicyLong (semantic-near, on-policy)} & \textbf{76.67} & -- \\
\bottomrule
\end{tabular}
\end{table}

\textbf{Difficulty matters.}
Replacing semantically similar hard negatives with \emph{bottom-$k$} (least similar) retrieval drops the average by $1.19$, consistent with prior findings that low-similarity negatives reduce training difficulty and weaken the contrastive signal. Replacing the secondary-retrieved negatives with fully \emph{random} chunks---i.e., not retrieving negatives from the verified positive contexts at all---is the weakest variant ($-2.90$), as it removes the contrastive structure entirely.

\textbf{Diversity is not the driver.}
For each query we additionally select a hard negative uniformly at random from the entropy candidate pool, which yields substantially \emph{higher} negative diversity than PolicyLong's nearest-neighbor selection. This higher-diversity variant is $2.47$ below PolicyLong, showing that the gain comes from aligning hard negatives with the current model's confusion frontier, not from increased diversity per se.

\textbf{The on-policy gain is orthogonal to the mining algorithm.}
In a paired experiment that mines hard negatives with BM25 instead of dense retrieval (the two rows differ \emph{only} in fixed vs.\ on-policy refresh), on-policy refresh raises the average from $74.01$ to $74.46$ ($+0.45$). Although BM25 is a comparatively weak hard-negative source here (both rows are well below the semantic-near setting), on-policy refresh still delivers a consistent positive contribution, indicating that the benefit of on-policy construction does not depend on the specific mining algorithm.

\end{document}

%% file: math_commands.tex
\usepackage{amsmath,amsfonts,bm}
\usepackage{multicol}
\def\eqref#1{equation~\ref{#1}}
\def\1{\bm{1}}

\DeclareMathAlphabet{\mathsfit}{\encodingdefault}{\sfdefault}{m}{sl}
\SetMathAlphabet{\mathsfit}{bold}{\encodingdefault}{\sfdefault}{bx}{n}

%% file: example_paper.bib
@article{fang2024wrong,
  title={What is Wrong with Perplexity for Long-context Language Modeling?},
  author={Fang, Lizhe and Wang, Yifei and Liu, Zhaoyang and Zhang, Chenheng and Jegelka, Stefanie and Gao, Jinyang and Ding, Bolin and Wang, Yisen},
  journal={arXiv preprint arXiv:2410.23771},
  year={2024}
}

@article{gao2024quest,
  title={{QUEST}: Query-Centric Data Synthesis Approach for Long-Context Scaling of Large Language Model},
  author={Gao, Chaochen and Wu, Xing and Fu, Qi and Hu, Songlin},
  journal={arXiv preprint arXiv:2405.19846},
  year={2024}
}

@article{nextlong2025,
  title={{NExtLong}: Toward Effective Long-Context Training without Long Documents},
  author={Gao, Chaochen and Wu, Xing and Lin, Zijia and Zhang, Debing and Hu, Songlin},
  journal={arXiv preprint arXiv:2501.12766},
  year={2025}
}

@misc{lozhkov2024fineweb-edu,
  title={Fineweb-edu: the finest collection of educational content, 2024},
  author={Lozhkov, Anton and Allal, Loubna Ben and von Werra, Leandro and Wolf, Thomas},
  journal={URL https://huggingface. co/datasets/HuggingFaceFW/fineweb-edu}
}

@misc{benallal2024cosmopedia,
  title={Cosmopedia, 2024},
  author={Allal, Loubna Ben and Lozhkov, Anton and Penedo, Guilherme and Wolf, Thomas and von Werra, Leandro},
  journal={URL https://huggingface. co/datasets/HuggingFaceTB/cosmopedia},
  volume={19}
}

@article{hsieh2024ruler,
  title={RULER: What's the Real Context Size of Your Long-Context Language Models?},
  author={Hsieh, Cheng-Ping and Sun, Simeng and Kriman, Samuel and Acharya, Shantanu and Rekesh, Dima and Jia, Fei and Zhang, Yang and Ginsburg, Boris},
  journal={arXiv preprint arXiv:2404.06654},
  year={2024}
}

@article{bai2024longbench2,
  title={Longbench v2: Towards deeper understanding and reasoning on realistic long-context multitasks},
  author={Bai, Yushi and Tu, Shangqing and Zhang, Jiajie and Peng, Hao and Wang, Xiaozhi and Lv, Xin and Cao, Shulin and Xu, Jiazheng and Hou, Lei and Dong, Yuxiao and others},
  journal={arXiv preprint arXiv:2412.15204},
  year={2024}
}

@article{su2021roformer,
  title={Roformer: Enhanced transformer with rotary position embedding},
  author={Su, Jianlin and Ahmed, Murtadha and Lu, Yu and Pan, Shengfeng and Bo, Wen and Liu, Yunfeng},
  journal={Neurocomputing},
  volume={568},
  pages={127063},
  year={2024},
  publisher={Elsevier}
}

@article{johnson2019billion,
  title={Billion-scale similarity search with GPUs},
  author={Johnson, Jeff and Douze, Matthijs and J{\'e}gou, Herv{\'e}},
  journal={IEEE Transactions on Big Data},
  volume={7},
  number={3},
  pages={535--547},
  year={2019},
  publisher={IEEE}
}

@article{dao2022flashattention,
  title={Flashattention: Fast and memory-efficient exact attention with io-awareness},
  author={Dao, Tri and Fu, Dan and Ermon, Stefano and Rudra, Atri and R{\'e}, Christopher},
  journal={Advances in neural information processing systems},
  volume={35},
  pages={16344--16359},
  year={2022}
}

@article{liu2023lost,
  title={Lost in the middle: How language models use long contexts},
  author={Liu, Nelson F and Lin, Kevin and Hewitt, John and Paranjape, Ashwin and Bevilacqua, Michele and Petroni, Fabio and Liang, Percy},
  journal={arXiv preprint arXiv:2307.03172},
  year={2023}
}

@article{settles2009active,
  title={Active learning literature survey},
  author={Settles, Burr},
  year={2009},
  publisher={University of Wisconsin-Madison Department of Computer Sciences}
}

@article{huo2025dots,
  title={dots. llm1 Technical Report},
  author={Huo, Bi and Tu, Bin and Qin, Cheng and Zheng, Da and Zhang, Debing and Zhang, Dongjie and Li, En and Guo, Fu and Yao, Jian and Lou, Jie and others},
  journal={arXiv preprint arXiv:2506.05767},
  year={2025}
}

@article{liu2024deepseek,
  title={Deepseek-v3 technical report},
  author={Liu, Aixin and Feng, Bei and Xue, Bing and Wang, Bingxuan and Wu, Bochao and Lu, Chengda and Zhao, Chenggang and Deng, Chengqi and Zhang, Chenyu and Ruan, Chong and others},
  journal={arXiv preprint arXiv:2412.19437},
  year={2024}
}

@article{sturua2024jina,
  title={jina-embeddings-v3: Multilingual embeddings with task lora},
  author={Sturua, Saba and Mohr, Isabelle and Akram, Mohammad Kalim and G{\"u}nther, Michael and Wang, Bo and Krimmel, Markus and Wang, Feng and Mastrapas, Georgios and Koukounas, Andreas and Wang, Nan and others},
  journal={arXiv preprint arXiv:2409.10173},
  year={2024}
}

@inproceedings{zhang2025re3syn,
  title={Re3Syn: A Dependency-Based Data Synthesis Framework for Long-Context Post-training},
  author={Zhang, Zhiyang and Liu, Ziqiang and Wang, Huiming and Shan, Renke and Kuang, Li and Wang, Lu and Soh, De Wen},
  booktitle={Proceedings of the 63rd Annual Meeting of the Association for Computational Linguistics (Volume 1: Long Papers)},
  pages={31468--31480},
  year={2025}
}

@article{xiong2023effective,
  title={Effective long-context scaling of foundation models},
  author={Xiong, Wenhan and Liu, Jingyu and Molybog, Igor and Zhang, Hejia and Bhargava, Prajjwal and Hou, Rui and Martin, Louis and Rungta, Rashi and Sankararaman, Karthik Abinav and Oguz, Barlas and others},
  journal={arXiv preprint arXiv:2309.16039},
  year={2023}
}

@inproceedings{
    shi2024incontext,
    title={In-Context Pretraining: Language Modeling Beyond Document Boundaries},
    author={Weijia Shi and Sewon Min and Maria Lomeli and Chunting Zhou and Margaret Li and Xi Victoria Lin and Noah A. Smith and Luke Zettlemoyer and Wen-tau Yih and Mike Lewis},
    booktitle={The Twelfth International Conference on Learning Representations},
    year={2024},
    url={https://openreview.net/forum?id=LXVswInHOo}
}

@article{jia2025entropylong,
  title={EntropyLong: Effective Long-Context Training via Predictive Uncertainty},
  author={Jia, Junlong and Chen, Ziyang and Wu, Xing and Gao, Chaochen and Lin, Zijia and Zhang, Debing and Hu, Songlin and Guo, Binghui},
  journal={arXiv preprint arXiv:2510.02330},
  year={2025}
}

@inproceedings{bengio2009curriculum,
  title={Curriculum learning},
  author={Bengio, Yoshua and Louradour, J{\'e}r{\^o}me and Collobert, Ronan and Weston, Jason},
  booktitle={Proceedings of the 26th annual international conference on machine learning},
  pages={41--48},
  year={2009}
}

@article{yen2024helmet,
  title={Helmet: How to evaluate long-context language models effectively and thoroughly},
  author={Yen, Howard and Gao, Tianyu and Hou, Minmin and Ding, Ke and Fleischer, Daniel and Izsak, Peter and Wasserblat, Moshe and Chen, Danqi},
  journal={arXiv preprint arXiv:2410.02694},
  year={2024}
}

@article{hubotter2026reinforcement,
  title={Reinforcement Learning via Self-Distillation},
  author={H{\"u}botter, Jonas and L{\"u}beck, Frederike and Behric, Lejs and Baumann, Anton and Bagatella, Marco and Marta, Daniel and Hakimi, Ido and Shenfeld, Idan and Buening, Thomas Kleine and Guestrin, Carlos and others},
  journal={arXiv preprint arXiv:2601.20802},
  year={2026}
}

@article{zhao2026self,
  title={Self-Distilled Reasoner: On-Policy Self-Distillation for Large Language Models},
  author={Zhao, Siyan and Xie, Zhihui and Liu, Mengchen and Huang, Jing and Pang, Guan and Chen, Feiyu and Grover, Aditya},
  journal={arXiv preprint arXiv:2601.18734},
  year={2026}
}

@article{qu2026pope,
  title={POPE: Learning to Reason on Hard Problems via Privileged On-Policy Exploration},
  author={Qu, Yuxiao and Setlur, Amrith and Smith, Virginia and Salakhutdinov, Ruslan and Kumar, Aviral},
  journal={arXiv preprint arXiv:2601.18779},
  year={2026}
}

@article{shenfeld2026self,
  title={Self-Distillation Enables Continual Learning},
  author={Shenfeld, Idan and Damani, Mehul and H{\"u}botter, Jonas and Agrawal, Pulkit},
  journal={arXiv preprint arXiv:2601.19897},
  year={2026}
}

@article{gao2025longmagpie,
  title={LongMagpie: A Self-synthesis Method for Generating Large-scale Long-context Instructions},
  author={Gao, Chaochen and Wu, Xing and Lin, Zijia and Zhang, Debing and Hu, Songlin},
  journal={arXiv preprint arXiv:2505.17134},
  year={2025}
}

@inproceedings{chen2020simclr,
  title={A Simple Framework for Contrastive Learning of Visual Representations},
  author={Chen, Ting and Kornblith, Simon and Norouzi, Mohammad and Hinton, Geoffrey},
  booktitle={Proceedings of the 37th International Conference on Machine Learning},
  pages={1597--1607},
  year={2020}
}

@inproceedings{robinson2020hard,
  title={Contrastive Learning with Hard Negative Samples},
  author={Robinson, Joshua and Chuang, Ching-Yao and Sra, Suvrit and Jegelka, Stefanie},
  booktitle={International Conference on Learning Representations},
  year={2021}
}

@inproceedings{karpukhin2020dense,
  title={Dense Passage Retrieval for Open-Domain Question Answering},
  author={Karpukhin, Vladimir and Oguz, Barlas and Min, Sewon and Lewis, Patrick and Wu, Ledell and Edunov, Sergey and Chen, Danqi and Yih, Wen-tau},
  booktitle={Proceedings of the 2020 Conference on Empirical Methods in Natural Language Processing},
  pages={6769--6781},
  year={2020}
}

@inproceedings{xiong2020ance,
  title={Approximate Nearest Neighbor Negative Contrastive Learning for Dense Text Retrieval},
  author={Xiong, Lee and Xiong, Chenyan and Li, Ye and Tang, Kwok-Fung and Liu, Jialin and Bennett, Paul and Ahmed, Junaid and Overwijk, Arnold},
  booktitle={International Conference on Learning Representations},
  year={2021}
}

@inproceedings{qu2021rocketqa,
  title={{RocketQA}: An Optimized Training Approach to Dense Passage Retrieval for Open-Domain Question Answering},
  author={Qu, Yingqi and Ding, Yuchen and Liu, Jing and Liu, Kai and Ren, Ruiyang and Zhao, Wayne Xin and Dong, Daxiang and Wu, Hua and Wang, Haifeng},
  booktitle={Proceedings of the 2021 Conference of the North American Chapter of the Association for Computational Linguistics},
  pages={5835--5847},
  year={2021}
}

@article{olmo2025olmo,
  title={Olmo 3},
  author={Olmo, Team and Ettinger, Allyson and Bertsch, Amanda and Kuehl, Bailey and Graham, David and Heineman, David and Groeneveld, Dirk and Brahman, Faeze and Timbers, Finbarr and Ivison, Hamish and others},
  journal={arXiv preprint arXiv:2512.13961},
  year={2025}
}

@inproceedings{jia2026litelong,
  title={LiteLong: Resource-Efficient Long-Context Data Synthesis for LLMs},
  author={Jia, Junlong and Wu, Xing and Gao, Chaochen and Chen, Ziyang and Lin, Zijia and Li, Zhongzhi and Wang, Weinong and Xu, Haotian and Jin, Donghui and Zhang, Debing and others},
  booktitle={Proceedings of the AAAI Conference on Artificial Intelligence},
  volume={40},
  number={37},
  pages={31274--31282},
  year={2026}
}
